\documentclass{article}

\usepackage{graphicx} 
\usepackage{wrapfig} 
\usepackage{subfigure} 
\usepackage{amsmath}
\usepackage{amsfonts}
\usepackage{natbib}

\usepackage{algorithm}
\usepackage{algorithmic}

\usepackage[accepted]{icml2012}

\icmltitlerunning{Disentangling Factors of Variation via Generative Entangling}

\def\citeA{\cite}

\begin{document} 

\twocolumn[
\icmltitle{Disentangling Factors of Variation via Generative Entangling}

\icmlauthor{Guillaume Desjardins}{desjagui@iro.umontreal.ca}
\icmlauthor{Aaron Courville}{courvila@iro.umontreal.ca}
\icmlauthor{Yoshua Bengio}{bengioy@iro.umontreal.ca} 
\icmladdress{D\'{e}partement d'informatique et de recherche op., Universit\'{e} de Montr\'{e}al
            Montr\'{e}al, QC H3C 3J7 CANADA}

\icmlkeywords{unsupervised learning, probabilistic model, undirected graphical model}

\vskip 0.3in
]

\begin{abstract} 
  Here we propose a novel model family with the objective of learning to
  disentangle the factors of variation in data. Our approach is based on
  the spike-and-slab restricted Boltzmann machine which we generalize to
  include higher-order interactions among multiple latent variables. Seen
  from a generative perspective, the multiplicative interactions emulates
  the entangling of factors of variation. Inference in the model can be
  seen as disentangling these generative factors. Unlike previous attempts
  at disentangling latent factors, the proposed model is trained using no
  supervised information regarding the latent factors.  We apply our model
  to the task of facial expression classification.
\end{abstract} 

\section{Introduction}
In many machine learning tasks, data originates from a generative process
involving complex interaction of multiple factors. Alone each factor
accounts for a source of variability in the data. Together their
interaction gives rise to the rich structure characteristic of many of the most
challenging domains of application.  Consider, for example, the task of
facial expression recognition. Two images of different individuals with the
same facial expression may result in images that are well separated in
pixel space. On the other hand, two images of the same individuals showing
different expressions may well be positioned very close together in pixel
space. In this simplified scenario, there are two factors at play: (1) the
identity of the individual, and (2) the facial expression. One of these
factors, the identity, is irrelevant to the task of facial expression
recognition and yet of the two factors it could well dominate the
representation of the image in pixel space. As a result, pixel space-based
facial expression recognition systems seem likely to suffer poor
performance due to the variation in appearance of individual faces.

Importantly, these interacting factors frequently do not combine as
simple superpositions that can be easily separated by choosing an
appropriate affine projection of the data. Rather, these factors often
appear tightly \emph{entangled} in the raw data. Our challenge is to construct
representations of the data that cope with the reality of entangled factors
of variation and provide features that may be appropriate to a wide variety of possible
tasks. In the context of our face data example, a representation capable of
disentangling identity and expression would be an effective representation
for either the facial recognition or facial expression classification.

In an effort to cope with these factors of variation, there has been a
broad-based movement in machine learning and in application domains such as
computer vision toward hand-engineering feature sets that are
\emph{invariant} to common sources of variation in data. This is the
motivation behind both the inclusion of feature pooling stages in the
convolutional network architecture \cite{LeCun89d} and the recent trend
toward representations based on large scale pooling of low-level features
~\cite{Wang2009,Coates2011-short}.  These approaches all stem from the powerful idea that 
invariant features of the data can be induced through the pooling together of a 
set of simple filter responses. Potentially even more powerful is the notion that one can actually \emph{learn} which filters to be pooled together from purely unsupervised data, and thereby extract directions of variance over
which the pooling features become invariant~\cite{Kohonen1979-short,Kohonen1996-short,Hyvarinen2000,Le2010-short,Koray-08,Ranzato2010b-short,Courville2011a}. However, in situations where there are multiple
relevant but entangled factors of variation that give rise to the data, we
require a means of feature extraction that \emph{disentangles} these
factors in the data rather than simply learn to represent some of these factors at the
expense of those that are lost in the filter pooling operation.

Here we propose a novel model family with the objective of \emph{learning
  to disentangle} the factors of variation evident in the data. Our
approach is based on the spike-and-slab restricted Boltzmann machine
(ssRBM) \cite{Courville+al-2011-small} which has recently been shown to be
a promising model of natural image data. We generalize the ssRBM to include
higher-order interactions among multiple binary latent variables. Seen from a generative perspective,
the multiplicative interactions of the binary latent variables emulates the
entangling of the factors that give rise to the data. Conversely, inference in the model
can be seen as an attempt to assign credit to the various
interacting factors for their combined account of the data -- in effect, to
disentangle the generative factors. Our approach relies only on unsupervised
approximate maximum likelihood learning of the model parameters, and as
such we do not require the use of any label information in defining the
factors to be disentangled. We believe this to be a research direction of
critical importance, as it is almost never the case that label information
exists for all factors responsible for variations in the data
distribution.

\section{Learning Invariant Features Versus Learning to Disentangle Features}

The principle that invariant features can actually \emph{emerge}, using
only unsupervised learning, from the organization of features into
subspaces was first established in the ASSOM model
\cite{Kohonen1996-short}. Since then, the same basic strategy has reappeared in a
number of different models and learning paradigms, including topological
independent component analysis~\cite{Hyvarinen2000,Le2010-short}, invariant
predictive sparse decomposition (IPSD) \cite{Koray-08}, as well as in
Boltzmann machine-based approaches~\cite{Ranzato2010b-short,Courville2011a}.  In
each case, the basic strategy is to group filters together by, for example,
using a variable (the pooling feature) that gates the activation for all
elements of the group. This gated activation mechanism causes the filters
within the group to share a common window on the dataset, which in turn leads
to filter groups composed of mutually complementary filters. In the end, the span
of the filter vectors defines a subspace which specifies the directions in
which the pooling feature is invariant.  Somewhat
surprisingly, this basic strategy has repeatedly demonstrated that useful
invariant features can be learned in a strictly unsupervised fashion, using
only the statistical structure inherent in the data. 
 While remarkable, one important problem with using this learning strategy 
is that the invariant representation formed by the pooling features offers a
somewhat incomplete view on the data as the detailed representation of
the lower-level features is abstracted away in the pooling procedure. 
While we would like higher level features to be more abstract and exhibit greater
invariance, we have little control over what information is lost through
feature subspace pooling. 

Invariant features,
by definition, have reduced sensitivity in the direction of
invariance. This is the goal of building invariant features and fully
desirable if the directions of invariance all reflect sources
of variance in the data that are uninformative to the task at hand.
However, it
is often the case that the goal of feature extraction is the
\emph{disentangling} or separation of
many distinct but informative factors in the data. In this situation, the
methods of generating invariant features -- namely, the feature subspace
method -- may be inadequate.
Returning to our facial expression classification example from the
introduction, consider a pooling feature made invariant to the expression
of a subject by forming a subspace of low-level filters that represent the
subject with various facial expressions (forming a basis for the subspace).
If this is the only pooling feature that is associated with the appearance
of this subject, then the facial expression information is lost to the
model representation formed by the set of pooling features. As illustrated in our
hypothetical facial expression classification task, this loss of information becomes a problem when the
information that is lost is necessary to successfully complete the task at
hand.

Obviously, what we really would like is for a particular feature set to be
invariant to the irrelevant features and disentangle the relevant
features. Unfortunately, it is often difficult to determine \emph{a priori}
which set of features will ultimately be relevant to the task at
hand. Further, as is often the case in the context of deep learning methods
\cite{CollobertR2008}, the feature
set being trained may be destined to be used in multiple tasks that may
have distinct subsets of relevant features.  Considerations such as these
lead us to the conclusion that the most robust approach to feature learning
is to disentangle as many factors as possible, discarding as little
information about the data as is practical. This is the motivation behind our proposed higher-order spike-and-slab Boltzmann machine.

\vspace{-1mm}
\section{Higher-order Spike-and-Slab Boltzmann Machines}
\vspace{-1mm}

\begin{figure*}[!htb]
\begin{align*}
E(v,s,f,g,h) \  =& \ 
    \frac{1}{2}v^{T} \Lambda v 
  - \sum_{k} e_{k}f_{k} + \sum_{i,k} c_{ik}g_{ik} + \sum_{j,k} d_{jk}h_{jk}
  + \frac{1}{2} \sum_{i,j,k} \alpha_{ijk}s_{ijk}^{2} \nonumber \\
& + \sum_{i,j,k} \left(
       - v^{T}W_{\cdot,ijk}s_{ijk}
       - \alpha_{ijk} \mu_{ijk}s_{ijk}
       + \frac{1}{2} \alpha_{ijk}\mu_{ijk}^{2}
    \right) g_{ik}h_{jk}f_{k},
\label{eq:energy}
\end{align*} \caption{Energy function of our higher-order spike \& slab RBM
(ssRBM), used to disentangle (multiplicative) factors of variation in the data.
Two groups of latent spike variables, $g$ and $h$, interact to explain the data
$v$, through the weight tensor $W$. While the ssRBM instantiates a slab
variable $s_j$ for each hidden unit $h_j$, our higher-order model employs a
slab $s_{ij}$ for each pair of spike variables ($g_i$,$h_j$).  $\mu_{ij}$ and
$\alpha_{ij}$ are respectively the mean and precision parameters of $s_{ij}$.
An additional set of spike variables $f$ are used to gate groups of latent
variables $h$, $g$ and serve to promote group sparsity. Most
parameters are thus indexed by an extra subscript $k$. Finally, $e$, $c$ and
$d$ are standard bias terms for variables $f$, $g$ and $h$, while $\Lambda$ is
a diagonal precision matrix on the visible vector.
\label{eq:energy}}
\end{figure*}

In this section, we introduce a model which makes some progress toward the ambitious goal of disentangling factors of variation. The model is based on the Boltzmann machine, an undirected graphical model. In particular we build on the spike-and-slab restricted Boltzmann Machine (ssRBM)~\cite{Courville2011a}, a model family that has previously shown promise as a means of learning invariant features via subspace pooling. 
The original ssRBM model possessed a limited form of higher-order interaction of two latent random variables: the spike and the slab
Our extension adds higher-order interactions between four distinct latent random variables. These include one set of slab variables and three interacting binary spike variables. Unlike the ssRBM, the interactions between the latent variables violate the conditional independence constraint of the restricted Boltzmann machine and therefore does not belong to this class of models. As a consequence, exact inference in the model is not tractable and we resort to a mean-field approximation.

Our strategy in promoting this model is that we intend to disentangle factors of variation via inference (recovering the posterior distribution over our latent variables) in a generative model. In the context of generative models, inference can roughly be thought of as running the generative process in reverse. Thus if we wish our inference process to disentangle factors of variation, our generative process should describe a means of factor entangling. The generative model we propose here represents one possible means of factor entangling. 

Let $v\in\mathbb{R}^{D}$ be the random visible vector that represents our observations with its 
mean zeroed. We build a latent representation of this data with binary latent variables $f\in \{0,1\}^{K}$, $g\in \{0,1\}^{M \times K}$ and $h\in
\{0,1\}^{N \times K}$. In the spike-and-slab context, we can think of $f$, $g$ and $h$ as a factored representation of the ``spike'' variables. We also include a set of real valued ``slab'' variables $s\in\mathbb{R}^{M \times N \times K}$, with element $s_{ijk}$ associated with hidden units $f_{k}$,
$g_{ik}$ and $h_{jk}$. The interaction between these variables is defined
through the energy function of Fig.~\ref{eq:energy}.

The parameters are defined as follows. $W\in\mathbb{R}^{D \times M \times N \times K}$ is a weight 4-tensor
connecting visible units to the interacting latent variables, these can be interpreted as forming a basis in image space;
$\mu\in\mathbb{R}^{M\times N \times K}$ and $\alpha\in\mathbb{R}^{M\times N \times K}$
are tensors describing the mean and precision of each $s_{ijk}$;
$\Lambda \in\mathbb{R}^{D\times D}$ is a diagonal precision matrix on the
visible vector;
and finally $c \in \mathbb{R}^{M \times K}$, $d \in \mathbb{R}^{N \times K}$ and $e \in \mathbb{R}^{K}$ are biases
on the matrices $g$, $h$ and vector $f$ respectively. The energy function 
fully specifies the joint probability distribution over the variables $v$, $s$
,$f$, $g$ and $h$: $p(v,s,f,g,h)=\frac{1}{Z}\exp\left\{ -E(v,s,f,g,h)\right\} $
where $Z$ is the partition function which ensures that the joint distribution is
normalized. 

\begin{figure}
    \centering
    \hspace{-3.0cm}
        \includegraphics[scale=0.25]{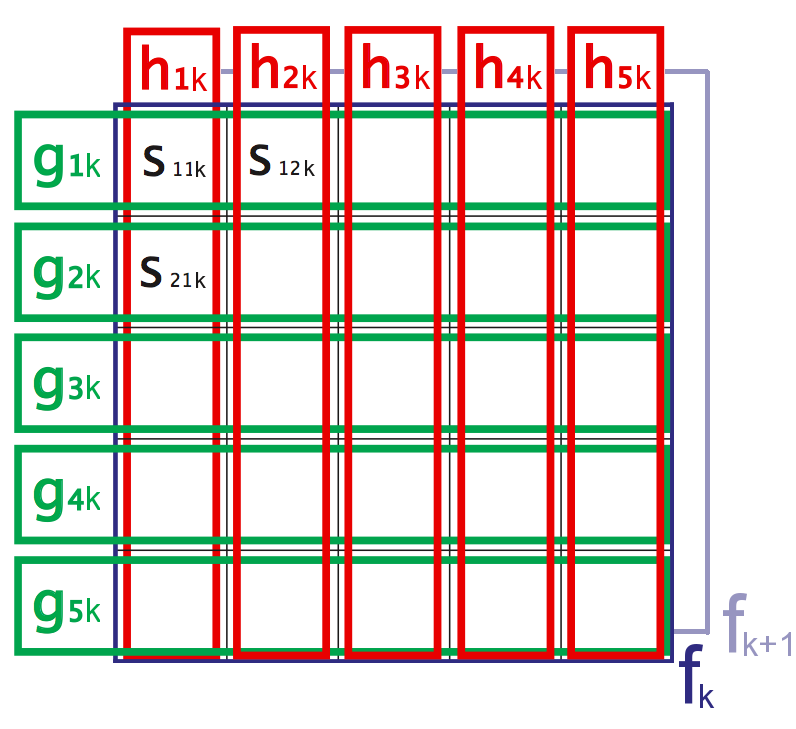}
    \hspace*{-3cm}
    \caption[]{\small{ 
    \label{fig:multipooling}
    Block-sparse connectivity pattern with dense interactions between $g$ and
    $h$ within each block (only shown for $k$-th block).  Each block is gated
    by a separate $f_k$ variable.
     }
    }
\end{figure}

As specified above, the energy function is similar to the ssRBM energy function \cite{Courville2011a,Courville+al-2011-small}, but includes a factored representation of the standard ssRBM spike variable. Yet, clearly the properties of the model are highly dependent on the topology
of the interactions between the real-valued
slab variables $s_{ijk}$, and three binary spike variables $f_{k}$, $g_{ik}$ and
$h_{jk}$.  We adopt a strategy that permits local interactions
within small groups of $f$, $g$ and $h$ in a block-like organizational pattern
as specified in Fig.~\ref{fig:multipooling}. The local block structure allows the model
to work incrementally towards disentangling the features by focusing on
manageable subparts of the problem. 

Similar to the standard spike-and-slab restricted Boltzmann machine \cite{Courville2011a,Courville+al-2011-small}, the energy function in Eq. \ref{eq:energy} gives rise to a Gaussian conditional over the visible variables:
\begin{equation}
p(v \mid s,f,g,h) = \mathcal{N}\left(\sum_{i,j,k} \Lambda^{-1}W_{\cdot ijk}s_{ijk}g_{ik}h_{jk}f_{k}\ ,\ \Lambda^{-1}\right)
\nonumber
\end{equation}
Here we have a four-way multiplicative interaction in the latent variables $s$, $f$, $g$ and $h$. The real-valued slab variable $s_{ijk}$ acts to scale the contribution of the weight vector $W_{\cdot ijk}$. As a consequence, after marginalizing out $s$, the factors $f$, $g$ and $h$ can also be seen as contributing both to the conditional mean and conditional variance of $p(v \mid f,g,h)$: 
\begin{align}
p(v \mid f,g,h) &= \mathcal{N}\left( \sum_{i,j,k} \Lambda^{-1}W_{\cdot ijk}\mu_{ijk}g_{ik}h_{jk}f_{k}, C_{v \mid f,g,h}\right) \nonumber \\
C_{v\mid f,g,h} &= \left(\Lambda - \sum_{i,j,k} W_{\cdot ijk}W_{\cdot ijk}^{T}\alpha_{ijk}^{-1}g_{ik}h_{jk}f_{k}\right)^{-1} \nonumber
\nonumber
\end{align}
This is an important property of the spike-and-slab framework that is also shared by other latent variable models of real-valued data such as the mean-covariance restricted Boltzmann machine (mcRBM) \cite{Ranzato2010b-short} and the mean Product of T-distributions model (mPoT) \cite{ranzato+mnih+hinton:2010-short}.

From a \emph{generative} perspective, the model can be thought of as consisting of
a set of $K$ factor blocks whose activity is gated by the $f$
variables. Within each block, the variables $g_{\cdot k}$ and $h_{\cdot k}$
can be thought of as local latent factors whose interaction gives rise to
the active block's contribution to the visible vector. Crucially, the
multiplicative interaction between the $g_{\cdot k}$ and $h_{\cdot k}$ for
a given block $k$ is mediated by the weight tensor $W_{\cdot, \cdot, \cdot,
  k}$ and the corresponding slab variables $s_{\cdot, \cdot, k}$. Contrary to more 
standard probabilistic factor models whose factors simply sum to give rise to the visible vector, 
the individual contributions of the elements of $g_{\cdot k}$ and $h_{\cdot k}$ are not easily isolated from one another. We can think of the generative process as \emph{entangling} the local block factor activations. 

From an \emph{encoding} perspective,  we are interested in using the posterior distribution over the latent variables as a representation or encoding of the data. Unlike in RBMs, in the case of the proposed model where we have higher-order interactions over the latent variables, the posterior over the latent variables does not factorize cleanly. By marginalizing over the slab variables $s$, we can recover a set of conditionals describing how the binary latent variables $f$, $g$ and $h$ interact. The conditional $P(f \mid v,g,h)$ is given below.
\begin{align}
P(f_{k} = 1\mid &v,g,h) = \mathrm{sigm} \left( c_{ik} + \sum_{i,j} v^{T}W_{\cdot ijk}\mu_{ijk}g_{ik}h_{jk} + \right. \nonumber \\
  &\left. \frac{1}{2} \sum_{i,j} \alpha_{ijk}^{-1}\left[v^{T}W_{\cdot ijk}\right]^{2}g_{ik}h_{jk} \right) 
\nonumber
\end{align}

It illustrates that with the factor configuration given in Fig.~\ref{fig:multipooling}, the factors $f_{k}$ are activated (assume value 1) through the sum-pooled response of all the weight vectors $W_{\cdot ijk}$ ($\forall 1 \leq i \leq M$ and $1 \leq j \leq N$) differentially gated by the values of $g_{ik}$  and $h_{jk}$, whose conditionals are respectively given by: 
\begin{align}
P(g_{ik} = 1 \mid &v,f, h)  =  \mathrm{sigm}\left(c_{ik} + \sum_{j}^{N}v^{T}W_{\cdot ijk}\mu_{ijk}h_{jk}f_{k} + \right.\nonumber \\
  &\left. \frac{1}{2} \sum_{j}^{N} \alpha_{ijk}^{-1} \left[v^{T}W_{\cdot ijk}\right]^{2}h_{jk}f_{k} \right) \nonumber \\
P(h_{jk} = 1\mid &v,f, g)  =  \mathrm{sigm}\left(d_{jk} + \sum_{i}^{M}v^{T}W_{\cdot ijk}\mu_{ijk}g_{ik}f_{k} + \right.\nonumber \\
 &\left. \frac{1}{2}\sum_{i}^{M}\alpha_{ijk}^{-1}\left[v^{T}W_{\cdot ijk}\right]^{2}g_{ik}f_{k} \right) \nonumber
\end{align}

For completeness, we also include the Gaussian conditional distribution over the slab variables $s$
\begin{equation}
p(s_{ijk}\mid v,f,g,h) = \mathcal{N}\left(\left[\alpha_{ijk}^{-1}v^{T}W_{\cdot ijk\cdot}+\mu_{ijk}\right]f_{k}g_{ik}h_{jk},\ \alpha_{ijk}^{-1}\right)
\nonumber
\end{equation}

From an encoding perspective, the gating pattern  on the $g$ and $h$ variables,
evident from Fig.~\ref{fig:multipooling} and from the conditionals distributions,
defines a form of local bilinear interaction \cite{tenenbaum00separating}. We
can interpret the values of $g_{ik}$ and $h_{jk}$ within block $k$ acting as
basis indicators, in dimensions $i$ and $j$, for the linear subspace in the
visible space defined by $W_{\cdot ijk} s_{ijk}$. 

From this perspective, we can think of $[g_{\cdot k}, h_{\cdot K}]$ as
defining a block-local binary coordination encoding of the data. Consider
the case illustrated by Fig.~\ref{fig:multipooling}, where we have $M=5$,
$N=5$ and the number of blocks ($K$) is 4. For each block, we have $M
\times N = 25$ filters which we encode using $M+N = 10$ binary latent
variables, where each $g_{ik}$ (alternately $h_{jk}$) effectively pools
over the subspace characterized by the variables $h_{jk}$, $1 \leq j \leq
N$ (alternately $g_{ik}$, $1 \leq i \leq M$) through their relative
interaction with $W_{\cdot ijk} s_{ijk}$. As a concrete example, imagine
that the structure of the weight tensor was such that, along the dimension
indexed by $i$, the weight vectors $W_{\cdot ijk}$ form oriented Gabor-like
edge detectors of different orientations. Yet along the dimension indexed
by $j$, the weight vectors $W_{\cdot ijk}$ form oriented Gabor-like edge
detectors of different colors. In this hypothetical example, $g_{ik}$
encodes orientation information while being invariance to the color of the
edge, while $h_{jk}$ encodes color information while being invariant to
orientation. Hence we could say that we have \emph{disentangled} the latent
factors.

\vspace{-1mm}
\subsection{Higher-order Interactions as a Multi-Way Pooling Strategy}
\vspace{-1mm}

As alluded to above, one interpretation of the role of $g$ and $h$ is as distinct and complementary sum-pooled feature sets. Returning to Fig.~\ref{fig:multipooling}, we can see that, for each block, the $g_{ik}$ pool across the columns of the $k$th block, along the $i$th row, while the $h_{\cdot k}$ pool across rows, along the $j$th column. The $f$ variables are also interpretable as pooling across all elements of the block. One way to interpret the complementary pooling structures of the $g$ and $h$ is as a multi-way pooling strategy. 

This particular pooling structure was chosen to study the potential of learning the kind of bilinear interaction that exists between the $g_{\cdot k}$ and $h_{\cdot k}$ within a block. The $f_{k}$ are present to promote block cohesion by gating the interaction of between $g_{\cdot k}$ and $h_{\cdot k}$ and the visible vector $v$.

This higher-order structure is of course just one choice of many possible
higher-order interaction architectures. One can easily imagine defining
arbitrary overlapping pooling regions, with the number of overlapping
pooling regions specifying the order of the latent variable interaction. We believe that explorations of overlapping pooling regions of this type is a promising direction of future inquiry. One potentially interesting direction is to consider overlapping blocks (such as our $f$ blocks). The overlap will define a topology over the features as they will share lower-level features (i.e. the slab variables). A topology thus defined could potentially be exploited to build higher-level data representations that possess local receptive fields. These kind of local receptive fields have been shown to be useful in building large and deep models that perform well in object classification tasks in natural images~\cite{Coates2011-short}.

\vspace{-1mm}
\subsection{Variational inference and unsupervised learning}
\vspace{-1mm}
Due to the multiplicative interaction between the latent variables $f$, $g$
and $h$, computation of {\small $P(f \mid v)$, $P(g \mid v)$}
and {\small $P(h \mid v)$} is intractable. While the slab variables also interact multiplicatively, 
we are able to analytically marginalize over them. Consequently we resort to a variational
approximation of the joint conditional {\small $P(f,g,h \mid v)$} with the standard
mean-field structure. i.e. we choose {\small $Q_{v}(f,g,h) = Q_{v}(f)Q_{v}(g)Q_{v}(h)$} 
such that the KL divergence
{\small $\mathrm{KL}(Q_{v}(f,g,h) \Vert P(f,g,h \mid v))$} is minimized, or
equivalently, that the variational lower bound {\small $\mathcal{L}(Q_{v})$} on the log
likelihood of the data is maximized: 
\begin{align}
\max_{Q_{v}} \mathcal{L}(Q_{v}) = \max_{Q_{v}} \sum_{f,g,h} & Q_{v}(f)Q_{v}(g)Q_{v}(h)\nonumber\\
&\log\left(\frac{p(f, g,h\mid v)}{Q_{v}(f)Q_{v}(g)Q_{v}(h)}\right), \nonumber
\end{align}
where the sums are taken over all values of the elements of $f$, $g$ and $h$
respectively. Maximizing this lower bound with respect to the variational
parameters $\hat{f}_k \equiv Q_{v}(f_{k} = 1)$, $\hat{g}_{ik} \equiv
Q_{v}(g_{ik} = 1)$ and $\hat{h}_{jk} \equiv Q_{v}(h_{jk} = 1)$, results in the
set of approximating factored distributions:
\begin{align}
\hat{f}_{k} = \mathrm{sigm} & \left(
     c_{ik} + \sum_{i,j} v^{T}W_{\cdot ijk}\mu_{ijk}\hat{g}_{ik}\hat{h}_{jk}\right. + \nonumber \\
     & \hspace{5mm} \left. \frac{1}{2} \sum_{i,j} \alpha_{ijk}^{-1}\left[v^{T}W_{\cdot ijk}\right]^{2}\hat{g}_{ik}\hat{h}_{jk} \right), \nonumber  \\
\hat{g}_{ik}  =  \mathrm{sigm} & \left(
    c_{ik} + \sum_{j}^{N}v^{T}W_{\cdot ijk}\mu_{ijk}\hat{h}_{jk}\hat{f}_{k}\right) \nonumber \\
    & \hspace{5mm} \left. + \frac{1}{2}\sum_{j}^{N}\alpha_{ijk}^{-1}\left[v^{T}W_{\cdot ijk}\right]^{2}\hat{h}_{jk}\hat{f}_{k}\right), \nonumber
\end{align}
\begin{align}
\hat{h}_{jk}  =  \mathrm{sigm} & \left(
    d_{jk} + \sum_{i}^{M}v^{T}W_{\cdot ijk}\mu_{ijk}\hat{g}_{ik}\hat{f}_{k}\right. \nonumber \\
    & \hspace{5mm} \left. + \frac{1}{2}\sum_{i}^{M}\alpha_{ijk}^{-1}\left[v^{T}W_{\cdot ijk}\right]^{2}\hat{g}_{ik}\hat{f}_{k}\right). \nonumber
\end{align}

The above equations form a set of fixed point equations which we iterate until the values of all $Q_v(f_{k})$, $Q_v(g_{ik})$ and $Q_v(h_{jk})$ converge. Since the expression for $\hat{f}_{k}$ does not depend on $\hat{f}_{k'}$, $\forall k'$, $\hat{g}_{ik}$ does not depend on $\hat{g}_{i'k'}$, $\forall i',k'$, and $\hat{h}_{jk}$ does not depend on $\hat{h}_{j'k'}$, $\forall i',k'$, we can define a three stage update strategy where we update the values of all $K$ values of $\hat{f}$ in parallel, then update all $K \times M$ values of $\hat{g}$ in parallel and finally update all $K \times N$ values of $\hat{h}$ in parallel. 

Following the variational EM training approach~\cite{Saul+96}, we alternately maximize the lower bound $\mathcal{L}(Q_{v})$ with respect to the variational parameters $\hat{f}$, $\hat{g}$ and $\hat{h}$ (E-step) and maximizing $\mathcal{L}(Q_{v})$ with respect to the model parameters (M-step). The gradient of $\mathcal{L}(Q_{v})$ with respect to the model parameters $\theta = \{ W, \mu, \alpha, \Lambda, b, c, d, e \}$  is given by:
\vspace{.5cm}
\begin{align}
&\frac{\partial\mathcal{L}(Q_{v})}{\partial\theta} = \sum_{f,g,h} Q_{v}(f)Q_{v}(g)Q_{v}(h)\mathbb{E}_{p(s\mid v,f,g,h)}\left[-\frac{\partial E}{\partial\theta}\right]\nonumber\\
&\hspace*{30mm}+\mathbb{E}_{p(v,s,g,h)}\left[\frac{\partial E}{\partial\theta}\right],
\nonumber
\label{eq:gradient}
\end{align}

where $E$ is the energy function given in Eq. \ref{eq:energy}. As is evident from Eq. \ref{eq:gradient}, the gradient of $\mathcal{L}(Q_{v})$ with respect to the model parameters contains two terms: a positive phase that depends on the data $v$ and a negative phase, derived from the partition function of the joint $p(v,s,f,g,h)$ that does not. We adopt a
training strategy similar to that of \cite{SalHinton09-short}, in that we combine a
variational approximation of the positive phase of the gradient with a block Gibbs
sampling-based stochastic approximation of the negative phase. Our Gibbs
sampler alternately samples, in parallel, each set of random variables,
sampling from $p(f \mid v, g,h)$, $p(g \mid v, f,h)$, $p(h \mid v, f, g)$,
$p(s \mid v,f,g,h)$, and finally sampling from $p(v \mid f,g,h,s)$.

\vspace{-1mm}
\subsection{The Challenge of Unsupervised Learning to Disentangle}
\vspace{-1mm}

Above we have briefly outline our procedure for training the unsupervised learning. The web of interactions between the latent random variables, particularly those between $g$ and $h$, makes the unsupervised learning of the model parameters a particularly challenging learning problem. It is the difficultly of learning that motivates our block-wise organization of the interactions between the $g$ and $h$ variables. The block structure allows the interactions between $g$ and $h$ to remain local, with each $g$ interacting with relatively few $h$ and each $h$ interacting with relatively few $g$. This local neighborhood structure allows the inference and learning procedures to better manage the complexities of teasing apart the latent variable interactions and adapting the model parameters to (approximately) maximize likelihood.

By using many of these blocks of local interactions we can leverage the known tractable learning properties of models such as the RBM.  
Specifically, if we consider each block as a kind of super hidden unit gated by $f$, then
with no interactions across blocks (apart from those mediated by the mutual connections to the visible units) the model assumes the form of an RBM. 

While our chosen interaction structure allows our higher-order model to be
able to learn, one consequence is that the model is only capable of
disentangling relatively local factors that appear within a single
block. We suggest that one promising avenue to accomplish more extensive disentangling is to consider stacking multiple version of the proposed model and consider layer-by-layer disentangling of the factors of variation present in the data.
The idea is to start with local disentangling and move gradually toward disentangling non local and more abstract factors. 

\vspace{-1mm}
\section{Related Work}
\vspace{-1mm}

The model proposed here was strongly influenced by previous attempts to
disentangle factors of variation in data using latent variable models. One
of the earlier efforts in this direction also used higher-order
interactions of latent variables, specifically
bilinear~\cite{tenenbaum00separating,Grimes2005short} and
multilinear~\cite{Vasilescu2005-short} models. 
One critical difference between these previous attempts to disentangle
factors of variation and our method is that unlike these previous
methods, we are attempting to learn to disentangle from entirely
unsupervised information. In this way, one can interpret our approach as an
attempt to extend the subspace feature pooling approach to the problem of
disentangling factors of variation.

Bilinear models are essentially
linear models where the higher-level state is factored into the product of two
variables. Formally, the elements
of observation $x$ are given by $x_k = \sum_i \sum_j
W_{ijk}y_iz_j$, $\forall k$, where $y_i$ and $z_j$ are elements of the two 
factors ($y$ and $z$) representing the observation and $W_{ijk}$ is an
element of the tensor of model parameters~\cite{tenenbaum00separating}. The tensor
$W$ can be thought of as a generalization of the typical weight matrix
found in most unsupervised models we have considered
above. \citeA{tenenbaum00separating} developed an EM-based algorithm to learn the
model parameters and demonstrated, using images of letters from a set of
distinct fonts, that the model could disentangle the style (font
characteristics) from content (letter identity).
\citeA{Grimes2005short} later
developed a bilinear sparse coding model of a similar form as described
above but included additional terms to the objective function to render the
elements of both $y$ and $z$ sparse. They also require observation of
the factors in order to train the model, and used the model to develop
transformation invariant features of natural images. Multilinear models are
simply a generalization of the bilinear model where the number of factors
that can be composed together is 2 or more. \citeA{Vasilescu2005-short} develop a
multilinear ICA model, which they use to model images of faces, to
disentangle factors of variation such as illumination, views (orientation
of the image plane relative to the face) and identities of the people. 

\citet{Hinton-transforming-aa-2011} also propose to disentangle factors
of variation by learning to extract features associated
with pose parameters, where the changes in pose parameters (but not
the feature values) are known at training time.
The proposed model is also closely related to recent work
\cite{Memisevic+Hinton-2010}, where higher-order Boltzmann Machines are
used as models of spatial transformations in images.  While there are a
number of differences between this model and ours, the most significant
difference is our use of multiplicative interactions between \emph{latent}
variables. While they included higher-order
interactions within the Boltzmann energy function, they were used
exclusively between observed variables, dramatically simplifying the
inference and learning procedures. Another major point of departure is that
instead of relying on low-rank approximations to the
weight tensor, our approach employs highly structured and sparse
connections between latent variables (e.g. $g_{ik}$ is not interact with or $h_{jk'}$ for $k' \neq k$), 
reminiscent of recent work on structured sparse coding \cite{gregor-nips-11} and structured
$l1$-norms \cite{Bach2011}.  As discussed above, our use of a sparse connection structure 
allows us to isolate groups of interacting latent variables. Keeping the interactions local 
in this way, is a key component of our ability to successfully learn using only unsupervised data.

\vspace{-1mm}
\section{Experiments}
\label{sec:exp}
\vspace{-1mm}

\subsection{Toy Experiment}

\begin{figure}
    \centering
    \subfigure
    {
        \label{fig:toy_data}
        \includegraphics[scale=1.5]{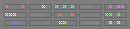}
    }
    \subfigure
    {
        \label{fig:toy_filters}
        \hspace{-4mm}
        \includegraphics[scale=1.9]{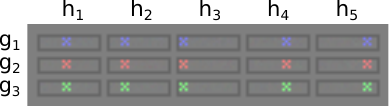}
    }
    \caption[]{(top) Samples from our synthetic dataset (before noise). In
    each image, a figure ``X'' can appear at five different positions, in one
    of eight basic colors. Objects in a given image must all be of the
    same color. (bottom) Filters learnt by a bilinear ssRBM with $M=3$, $N=5$,
    which succesfully show disentangling of color information (rows) from
    position (columns).
    \small{\label{fig:toy_exp}}}
\end{figure}

We showcase the ability of our model to disentangle factors of variation, by
training it on a synthetic dataset, a subset of which is shown in
Fig.~\ref{fig:toy_exp}~(top). Each color image, of size $3\times20$ is composed of one
basic object of varying color, which can appear at five different
positions. The constraint is that all objects in a given image must be of the
same color. Additive gaussian noise is super-imposed on the resulting images to
facilitate mixing of the RBM negative phase.
A bilinear ssRBM with $M=3$ and $N=5$ should in theory have the capacity
to disentangle the two factors of variation present in the data, as there are
$2^3$ possible colors and $2^5$ configurations of object placement.
The resulting filters are shown in Fig.~\ref{fig:toy_exp}~(bottom): the
model has succesfully learnt a binary encoding of color along $g$-units (rows)
and positions along $h$ (columns). Note that this would have been
extremely difficult to perform without multiplicative interactions of latent
variables: an RBM with $15$ hidden units technically has the capacity to learn
similar filters, however it would be incapable of enforcing mutual exclusivity
between hidden units of different color. The bilinear ssRBM model on the other
hand generates near-perfect samples (not shown), while factoring the
representation for use in deeper layers.

\subsection{Toronto Face Dataset}

We evaluate our model on the recently introduced Toronto Face Dataset (TFD)
\cite{Susskind2010}, which contains a large number of black \& white
$48\times48$ preprocessed facial images. These span a wide range of identities
and emotions and as such, the dataset is well suited to study the problem of
disentangling: models which can successfully separate identity from emotion
should perform well at the supervised learning task, which involves classifying
images into one of seven categories: \{anger, disgust, fear, happy, sad,
surprise, neutral\}. The dataset is divided into two parts: a large unlabeled
set (meant for unsupervised feature learning) and a smaller labeled set.  Note
that emotions appear much more prominently in the latter, since these are acted
out and thus prone to exaggeration. In contrast, most of the unlabeled set
contains natural expressions over a wider range of individuals.

In the course of this work, we have made several key refinements to the
original spike-and-slab formulation. Notably, since the slab variables
$\{s_{ijk}; \forall j\}$ can be interpreted as coordinates in the subspace of
the spike variable $g_{ik}$ (which spans the set of filters $\{W_{\cdot,ijk},
\forall j\}$), it is natural for these filters to be unit-norm. Each maximum
likelihood gradient update is thus followed by a projection of the filters onto
the unit-norm ball. Similarly, there exists an over-parametrization in the
direction of $W_{\cdot,ijk}$ and the sign of $\mu_{ijk}$, the parameter
controlling the mean of $s_{ijk}$. We thus constrain $\mu_{ijk}$ to be
positive, in our case greater than 1. Similar constraints are applied
on $B$ and $\alpha$ to ensure that the variances on the visible and slab
variables remain bounded.
While previous work \cite{Courville+al-2011-small} used the expected value of
the spike variables as the input to classifiers, or higher-layers in deep
networks, we found that the above re-parametrization consistently lead to
better results when using the product of expectations of $h$ and $s$. For
pooled models, we simply take the product of each binary spike, with the norm
of its associated slab vector. 

\vspace{-1mm}
\paragraph{Disentangling Emotion from Identity.}
\vspace{-1mm}
\label{sec:exp1}

\begin{figure}
    \centering
    \subfigure
    {
        \label{fig:results1}
        \includegraphics[scale=0.15]{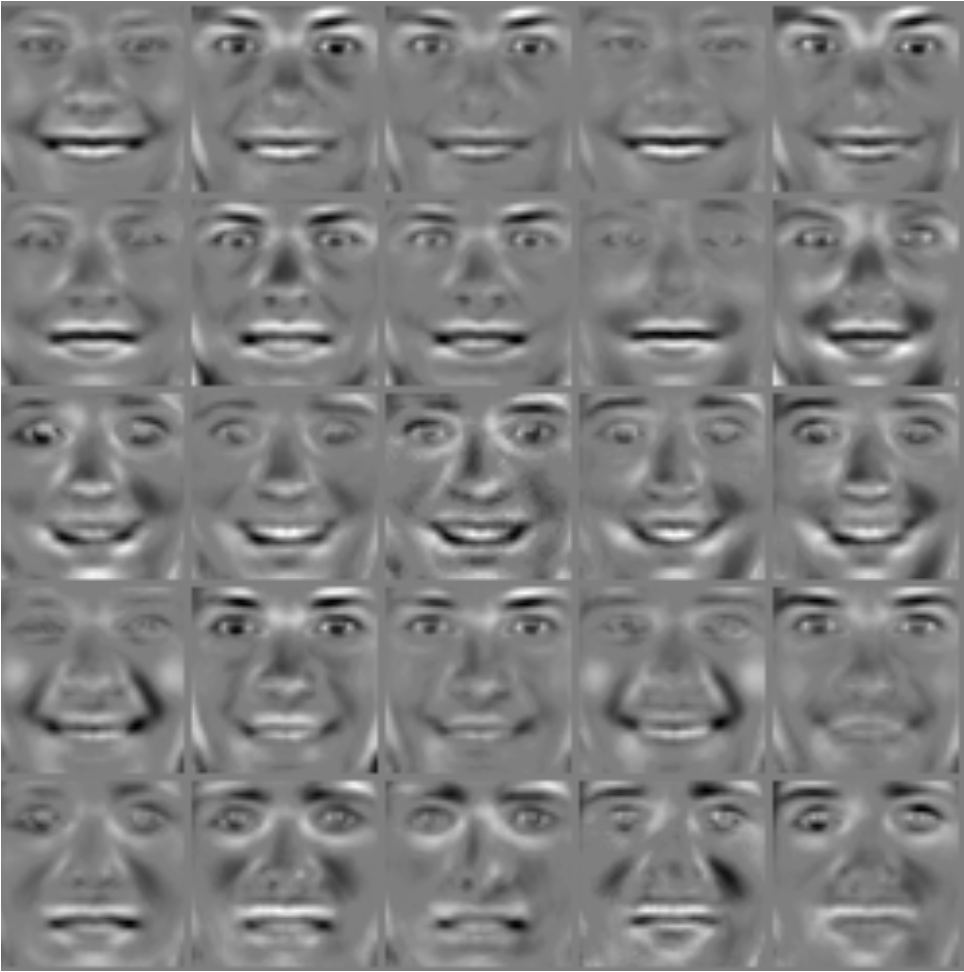}
    }
    \subfigure
    {
        \label{fig:results2}
        \includegraphics[scale=0.15]{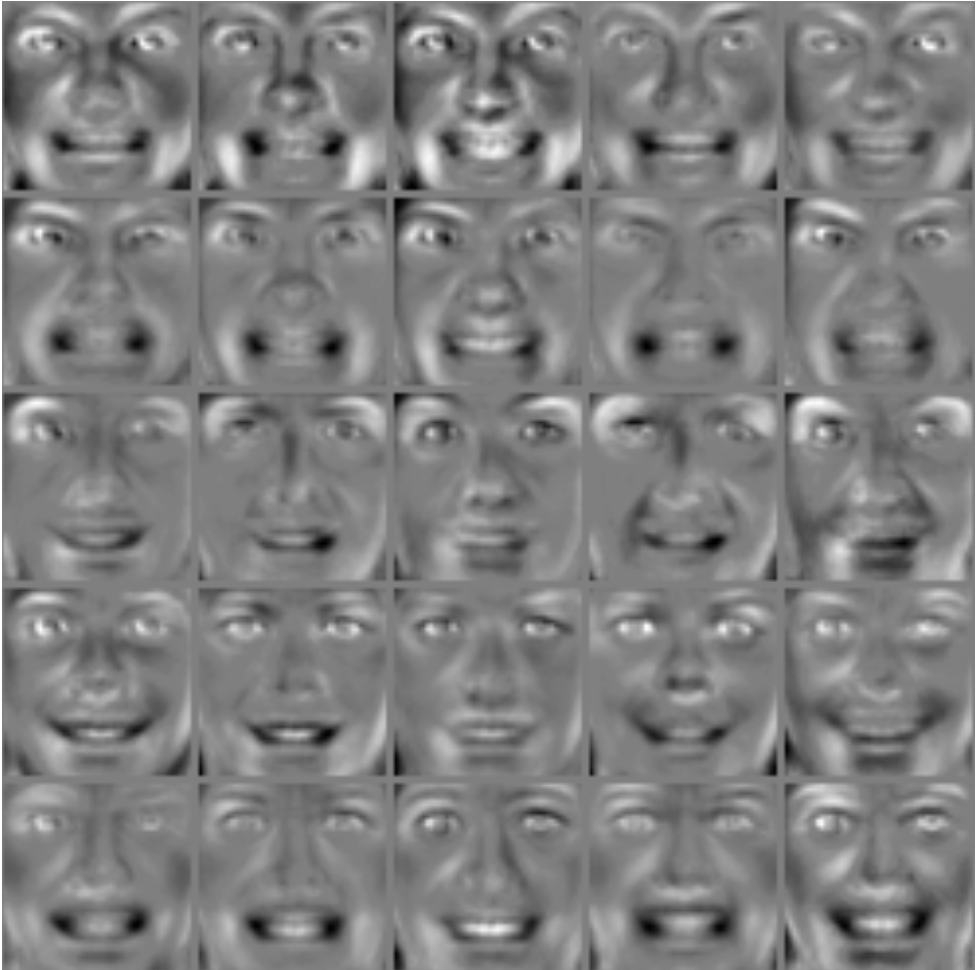}
    }
    \caption[]{\small{ 
    Example blocks obtained with $K=100$, $M=N=5$. The filters (inner-most
    dimension of tensor $W$) in each block exhibit global cohesion,
    specializing themselves to a subset of identities and emotions:
    \{happiness, fear, neutral\} in (left) and \{happiness, anger\} in (right). In
    both cases, $g$-units (which pool over columns) encode emotions, while
    $h$-units (which pool over rows) are more closely tied to identity.}
    \label{fig:filters}}
\end{figure}

We begin with a qualitative evaluation of our model, by visualizing the learned
filters (inner-most dimension of the matrix $W$) and pooling structures.  We
trained a model with $K=100$ and $M=N=5$ (that is to say $100$ blocks of
$5\times5$ interacting $g$ and $h$ units) on a weighted combination of the labeled and
unlabeled training sets. Doing so (as opposed to training on the unlabeled set
only) allows for greater interpretability of the results, as emotion is a more
prominent factor of variation in the labeled set). The results, shown in
Figure~\ref{fig:filters}, clearly show global cohesion within blocks pooled by
$f_k$, with row and column structure correlating with variances in
appearance/identity and emotions.

\vspace{-1mm}
\paragraph{Disentangling via Unsupervised Feature Learning.}
\vspace{-1mm}
\label{sec:exp2}

We now evaluate the representation learnt by our disentangling RBM, by
measuring its usefulness for the task of emotion recognition.  Our main
objective here is to evaluate the usefulness of disentangling, over traditional
approaches of pooling, as well as the use of larger, unpooled models. We thus
consider ssRBMs with $3000$ and $5000$ features, with either 
(i) no pooling (i.e. $K=5000$ spikes with $N=1$ slabs per spike),
(ii) pooling along a single dimension (i.e. $K=1000$ spike variables, pooling
$N=5$ slabs) or (iii) disentangled through our higher-order ssRBM (i.e.
$K=200$, with $g$ and $h$ units arranged in a $M \times N$ grid, with $M=N=5$).

We followed the standard TFD training protocol of performing unsupervised
training on the unlabeled set, and then using the learnt representation as
input to a linear SVM, trained and cross-validated on the labeled set.
Table~\ref{tab:results} shows the test accuracy obtained by various spike-and-slab
models, averaged over the 5-folds. 



\begin{table*}
\small
\begin{center}
\begin{tabular}{llllcccc}
             &        &         &         & \multicolumn{2}{c}{Factored} & \multicolumn{2}{c}{Unfactored} \\
{\bf Model}  &{\bf K} & {\bf M} & {\bf N} & valid & test              & valid & test \\
\hline \\
ssRBM   & 3000 &   & 1 & n/a & n/a  & 76.0\%  & 75.7\% \\ 
ssRBM   & 999  &   & 3 & 72.9\% & 74.4\% & 74.9\% & 73.5\% \\
hossRBM & 330  & 3 & 3 & 76.0\% & 75.7\% & 75.3\% & 75.2\% \\
hossRBM & 120  & 5 & 5 & 71.4\% & 70.7\% & 74.5\% & 74.2\% \\
\hline \\
ss-RBM  & 5000 &   & 1 & n/a & n/a & 76.7\% & 76.3\% \\
ss-RBM  & 1000 &   & 5 & 74.2\% & 74.0\% & 75.9\% & 74.6\% \\
hossRBM & 555 & 3 & 3 & 77.6\% & {\bf 77.4\%} & 76.2\% & 75.9\% \\
hossRBM & 200 & 5 & 5 & 73.3\% & 73.3\% & 75.6\% & 75.3\% \\
\end{tabular}\\
\end{center}
\caption{Classification accuracy for Toronto Face Dataset. We compare our
higher-order ssRBM for various block sizes $K$ and pooling regions $M\times N$.
The comparison is against first-order ssRBMs, which thus pool in a single
dimension of size $N$. First four models contain approximately $3,000$ filters,
while bottom four contain $5,000$. In both cases, we compare the effect of
using the factored representation, to the unfactored representation.}
\label{tab:results}
\end{table*}

We report two sets of numbers for models with pooling or disentangling: one
where we use the ``factored representation'', which is the element-wise product
of spike variables with the norm of their associated slab vector, and the
``unfactored representation'': the higher-dimensional representation formed by
considering all slab variables, each multiplied by their associated spikes.

We can see that the higher-order ssRBM achieves the best result: $77.4\%$,
using the factored representation. The fact that that our model outperforms the
``unfactored'' one, confirms our disentangling hypothesis: our model has
successfully learnt a lower-dimensional (factored) representation of the data,
useful for classification.
For reference, a linear SVM classifier on the pixels achieves $71.5\%$
\cite{Susskind2010}, an MLP trained with supervised backprop $72.72\%$\footnote{Salah Rifai, personal communication.},
while a deep mPoT model \cite{Ranzato2011-short}, which exploits local receptive
fields achieves $82.4\%$. 


\vspace{-1mm}
\section{Conclusion}
\vspace{-1mm}
We have presented a higher-order extension of the spike-and-slab restricted Boltzmann machine that factors the standard binary spike variable into three interacting factors. From a generative perspective, these interactions act to entangle the factors represented by the latent binary variables. Inference is interpreted as a process of disentangling the factors of variation in the data. As previously mentioned, we believe an important direction of future research to be the exploration of methods to gradually disentangle the factors of variation by stacking multiple instantiations of proposed model into a deep architecture. 

\small
\bibliography{strings,strings-shorter,ml,aigaion}

\begin{thebibliography}{24}
\providecommand{\natexlab}[1]{#1}
\providecommand{\url}[1]{\texttt{#1}}
\expandafter\ifx\csname urlstyle\endcsname\relax
  \providecommand{\doi}[1]{doi: #1}\else
  \providecommand{\doi}{doi: \begingroup \urlstyle{rm}\Url}\fi

\bibitem[Bach et~al.(2011)Bach, Jenatton, Mairal, and Obozinski]{Bach2011}
F.~Bach, R.~Jenatton, J.~Mairal, and G.~Obozinski.
\newblock Structured sparsity through convex optimization.
\newblock \emph{CoRR}, abs/1109.2397, 2011.

\bibitem[Coates et~al.(2011)Coates, Lee, and Ng]{Coates2011-short}
A.~Coates, H.~Lee, and A.~Y. Ng.
\newblock An analysis of single-layer networks in unsupervised feature
  learning.
\newblock In \emph{Proceedings of the Thirteenth International Conference on
  Artificial Intelligence and Statistics (AISTATS 2011)}, 2011.

\bibitem[Collobert and Weston(2008)]{CollobertR2008}
R.~Collobert and J.~Weston.
\newblock A unified architecture for natural language processing: Deep neural
  networks with multitask learning.
\newblock In W.~W. Cohen, A.~McCallum, and S.~T. Roweis, editors, \emph{{ICML}
  2008}, pages 160--167. ACM, 2008.

\bibitem[Courville et~al.(2011{\natexlab{a}})Courville, Bergstra, and
  Bengio]{Courville+al-2011-small}
A.~Courville, J.~Bergstra, and Y.~Bengio.
\newblock Unsupervised models of images by spike-and-slab {RBM}s.
\newblock In \emph{ICML'2011}, 2011{\natexlab{a}}.

\bibitem[Courville et~al.(2011{\natexlab{b}})Courville, Bergstra, and
  Bengio]{Courville2011a}
A.~Courville, J.~Bergstra, and Y.~Bengio.
\newblock A {S}pike and {S}lab {R}estricted {B}oltzmann {M}achine.
\newblock In \emph{AISTATS'2011}, 2011{\natexlab{b}}.

\bibitem[Gregor et~al.(2011)Gregor, Szlam, and LeCun]{gregor-nips-11}
K.~Gregor, A.~Szlam, and Y.~LeCun.
\newblock Structured sparse coding via lateral inhibition.
\newblock In \emph{Advances in Neural Information Processing Systems (NIPS
  2011)}, volume~24, 2011.

\bibitem[Grimes and Rao(2005)]{Grimes2005short}
D.~B. Grimes and R.~P. Rao.
\newblock {Bilinear sparse coding for invariant vision.}
\newblock \emph{Neural computation}, 17\penalty0 (1):\penalty0 47--73, January
  2005.

\bibitem[Hinton et~al.(2011)Hinton, Krizhevsky, and
  Wang]{Hinton-transforming-aa-2011}
G.~Hinton, A.~Krizhevsky, and S.~Wang.
\newblock Transforming auto-encoders.
\newblock In \emph{ICANN'2011: International Conference on Artificial Neural
  Networks}, 2011.

\bibitem[Hyv{\"{a}}rinen and Hoyer(2000)]{Hyvarinen2000}
A.~Hyv{\"{a}}rinen and P.~Hoyer.
\newblock Emergence of phase and shift invariant features by decomposition of
  natural images into independent feature subspaces.
\newblock \emph{Neural Computation}, 12\penalty0 (7):\penalty0 1705--1720,
  2000.

\bibitem[Kavukcuoglu et~al.(2009)Kavukcuoglu, Ranzato, Fergus, and
  {LeCun}]{Koray-08}
K.~Kavukcuoglu, M.~Ranzato, R.~Fergus, and Y.~{LeCun}.
\newblock Learning invariant features through topographic filter maps.
\newblock In \emph{Proceedings of the Computer Vision and Pattern Recognition
  Conference (CVPR'09)}, pages 1605--1612. IEEE, 2009.

\bibitem[Kohonen(1996)]{Kohonen1996-short}
T.~Kohonen.
\newblock Emergence of invariant-feature detectors in the adaptive-subspace
  self-organizing map.
\newblock \emph{Biological Cybernetics}, 75:\penalty0 281--291, 1996.
\newblock ISSN 0340-1200.

\bibitem[Kohonen et~al.(1979)Kohonen, Nemeth, Bry, Jalanko, and
  Riittinen]{Kohonen1979-short}
T.~Kohonen, G.~Nemeth, K.-J. Bry, M.~Jalanko, and H.~Riittinen.
\newblock Spectral classification of phonemes by learning subspaces.
\newblock In \emph{ICASSP '79.}, volume~4, pages 97 -- 100, 1979.

\bibitem[Le et~al.(2010)Le, Ngiam, Chen, hao Chia, Koh, and Ng]{Le2010-short}
Q.~Le, J.~Ngiam, Z.~Chen, D.~J. hao Chia, P.~W. Koh, and A.~Ng.
\newblock Tiled convolutional neural networks.
\newblock In \emph{NIPS'2010}, 2010.

\bibitem[{LeCun} et~al.(1989){LeCun}, Jackel, Boser, Denker, Graf, Guyon,
  Henderson, Howard, and Hubbard]{LeCun89d}
Y.~{LeCun}, L.~D. Jackel, B.~Boser, J.~S. Denker, H.~P. Graf, I.~Guyon,
  D.~Henderson, R.~E. Howard, and W.~Hubbard.
\newblock Handwritten digit recognition: Applications of neural network chips
  and automatic learning.
\newblock \emph{IEEE Communications Magazine}, 27\penalty0 (11):\penalty0
  41--46, Nov. 1989.

\bibitem[Memisevic and Hinton(2010)]{Memisevic+Hinton-2010}
R.~Memisevic and G.~E. Hinton.
\newblock Learning to represent spatial transformations with factored
  higher-order boltzmann machines.
\newblock \emph{Neural Computation}, 22\penalty0 (6):\penalty0 1473--1492, June
  2010.

\bibitem[Ranzato and Hinton(2010)]{Ranzato2010b-short}
M.~Ranzato and G.~H. Hinton.
\newblock Modeling pixel means and covariances using factorized third-order
  {B}oltzmann machines.
\newblock In \emph{CVPR'2010}, pages 2551--2558, 2010.

\bibitem[Ranzato et~al.(2010)Ranzato, Mnih, and
  Hinton]{ranzato+mnih+hinton:2010-short}
M.~Ranzato, V.~Mnih, and G.~Hinton.
\newblock Generating more realistic images using gated {MRF}'s.
\newblock In \emph{NIPS'2010}, 2010.

\bibitem[Ranzato et~al.(2011)Ranzato, Susskind, Mnih, and
  Hinton]{Ranzato2011-short}
M.~Ranzato, J.~Susskind, V.~Mnih, and G.~Hinton.
\newblock On deep generative models with applications to recognition.
\newblock In \emph{CVPR'2011}, 2011.

\bibitem[Salakhutdinov and Hinton(2009)]{SalHinton09-short}
R.~Salakhutdinov and G.~Hinton.
\newblock Deep {B}oltzmann machines.
\newblock In \emph{AISTATS'2009}, volume~5, pages 448--455, 2009.

\bibitem[Saul et~al.(1996)Saul, Jaakkola, and Jordan]{Saul+96}
L.~K. Saul, T.~Jaakkola, and M.~I. Jordan.
\newblock Mean field theory for sigmoid belief networks.
\newblock \emph{Journal of Artificial Intelligence Research}, 4:\penalty0
  61--76, 1996.

\bibitem[Susskind et~al.(2010)Susskind, Anderson, and Hinton]{Susskind2010}
J.~Susskind, A.~Anderson, and G.~E. Hinton.
\newblock The {T}oronto face dataset.
\newblock Technical Report UTML TR 2010-001, U. Toronto, 2010.

\bibitem[Tenenbaum and Freeman(2000)]{tenenbaum00separating}
J.~B. Tenenbaum and W.~T. Freeman.
\newblock Separating style and content with bilinear models.
\newblock \emph{Neural Computation}, 12\penalty0 (6):\penalty0 1247--1283,
  2000.

\bibitem[Vasilescu and Terzopoulos(2005)]{Vasilescu2005-short}
M.~A.~O. Vasilescu and D.~Terzopoulos.
\newblock {Multilinear independent components analysis}.
\newblock In \emph{CVPR'2005}, volume~1, pages 547--553, 2005.

\bibitem[Wang et~al.(2009)Wang, Ullah, Kl\"aser, Laptev, and Schmid]{Wang2009}
H.~Wang, M.~M. Ullah, A.~Kl\"aser, I.~Laptev, and C.~Schmid.
\newblock Evaluation of local spatio-temporal features for action recognition.
\newblock In \emph{British Machine Vision Conference (BMVC)}, pages 127--127,
  London, UK, September 2009.

\end{thebibliography}
\bibliographystyle{abbrvnat}

\end{document}